\documentclass[a4paper,fleqn]{cas-dc}
\usepackage{adjustbox}
  
\usepackage{amsmath,amssymb}
\usepackage{multirow}
\usepackage{bbm}
\usepackage{booktabs}
\usepackage[linesnumbered, ruled]{algorithm2e}
\usepackage{color}
\makeatletter       
\DeclareRobustCommand\onedot{\futurelet\@let@token\@onedot}
\def\@onedot{\ifx\@let@token.\else.\null\fi\xspace}
\def\eg{\emph{e.g}\onedot} 
\def\ie{\emph{i.e}\onedot}

\def\wrt{w.r.t\onedot}

\newcommand{\highlight}[1]{\textcolor{black}{#1}}
\hypersetup{
    colorlinks=true,
    linkcolor=cyan,
    filecolor=cyan,      
    urlcolor=cyan,
    citecolor=cyan,
}
   
\usepackage[numbers]{natbib}
\def\tsc#1{\csdef{#1}{\textsc{\lowercase{#1}}\xspace}}
\tsc{WGM}
\tsc{QE}
\begin{document}
\let\WriteBookmarks\relax
\def\floatpagepagefraction{1}
\def\textpagefraction{.001}
  
% \shorttitle{3D Object Detection for Autonomous Driving: A Survey}    

\shortauthors{R. Qian, X. Lai and X. Li}  

\title [mode = title]{BADet: Boundary-Aware 3D Object Detection from Point Clouds}  

\author[1]{Rui Qian}
\ead{qiianruii@gmail.com} 

\address[1]{Key Lab of Data Engineering and Knowledge Engineering, 
Renmin University of China, Beijing 100872, China.}

\author[2]{Xin Lai}
\ead{laixin@ruc.edu.cn} 
\address[2]{School of Mathematics, Renmin University of China, Beijing 100872, China}

\author[1]{Xirong Li}
\cormark[1] 
% \fnmark[1,3]
\ead{xirong@ruc.edu.cn} 
   
% \address[3]{\TeX{} Users Group, Providence, MA, USA}

\cortext[cor1]{Corresponding author} 

% \cortext[cor2]{Principal corresponding author} 
\nonumnote{
\href{https://doi.org/10.1016/j.patcog.2022.108524}{https://doi.org/10.1016/j.patcog.2022.108524}}
\nonumnote{
  \text{©} 2022 Elsevier Ltd. All rights reserved.}

\begin{abstract}
    Currently, existing state-of-the-art 3D object detectors are in two-stage paradigm. These 
    methods typically comprise two steps: 1) Utilize a region proposal network to propose a handful 
    of high-quality proposals in a bottom-up fashion. 2) Resize and pool the semantic features from
    the proposed regions to summarize RoI-wise representations for further refinement. Note that 
    these RoI-wise representations in step 2) are considered individually as uncorrelated entries when fed 
    to following detection headers. Nevertheless, we observe these proposals generated by step 1) offset 
    from ground truth somehow, emerging in local neighborhood densely with an underlying probability. 
    Challenges arise in the case where a proposal largely forsakes its boundary information due to 
    coordinate offset while existing networks lack corresponding information compensation mechanism. 
    In this paper, we propose \emph{BADet} for 3D object detection from point clouds. Specifically, instead 
    of refining each proposal independently as previous works do, we represent each proposal as a node 
    for graph construction within a given cut-off threshold, associating proposals in the form of 
    local neighborhood graph, with boundary correlations of an object being explicitly exploited. Besides, 
    we devise a lightweight \emph{Region Feature Aggregation Module} to fully exploit voxel-wise, pixel-wise, 
    and point-wise features with expanding receptive fields for more informative RoI-wise representations. 
    \highlight{We validate BADet both on widely used KITTI Dataset and highly challenging nuScenes Dataset.}
    As of Apr. 17th, 2021, our BADet achieves on par performance on KITTI 3D detection leaderboard and 
    ranks $1^{st}$ on $Moderate$ difficulty of $Car$ category on KITTI BEV detection leaderboard. 
    The source code is available at \href{https://github.com/rui-qian/BADet}{\emph{https://github.com/rui-qian/BADet}}.
    
    \textcolor{white}{\text{©} 2022 Elsevier Ltd. All rights reserved.}
       
    \rightline{\text{©} 2022 Elsevier Ltd. All rights reserved.}
\end{abstract}
  

\maketitle
\section{INTRODUCTION} \label{sec:intro}

% \IEEEPARstart{A}UTONOMOUS driving has the potential to radically change people's lives, improving mobility and reducing travel time, energy consumption, and emissions. Therefore, unsurprisingly, in the last decade both research and industry have put significant efforts to develop self-driving vehicles. As one of the key enabling technologies for autonomous driving, 3D object detection has received a lot of attention. In particular, lately, deep learning-based 3D object detection approaches have gained popularity.

% Existing 3D object detection approaches can be roughly categorized into two groups according to whether the input data are images or LiDAR signals (generally represented as point clouds). Compared with the LiDAR-based methods, approaches which estimate the 3D bounding boxes from images only, face a much greater challenge, as recovering 3D information from 2D input data is an ill-posed problem. However, despite this intrinsic difficulty, over the past six years, image-based 3D object detection methods have proliferated in the computer vision (CV) community. More than 80 papers have been published on the top-tier conferences and journals in this area, achieving several breakthroughs both in terms of detection accuracy and inference speed.

% \highlight{
Scene understanding has been a long-term interest of pattern recognition \cite{WANG2021107884,KIM2021108068, 
CUPEC2020107199,LIANG2021107630}. 3D object detection, as a key step towards scene understanding in the real world, 
is to estimate 3D bounding boxes of real objects from sensory data. Point clouds, which effectively capture 
informative geometric attributes such as 3D positions, orientations and occupied volumes, are increasingly 
prevalent sensor data, in particular for 3D object detection in driving scenes \cite{he2020sassd,pvrcnn2020,ciassd}. 
Despite existing efforts \cite{rahman2019recent, arnold2019survey, guo2019deep}, 3D object detection has still 
trailed 2D counterparts thus far \cite{girshickICCV15fastrcnn,RenHGS15fasterrcnn,LinGGHD17RetinaNet}. 
This paper is targeted at detecting 3D objects from point clouds. 
% }
       
\highlight{
Depending on how feature representation learning is performed on point clouds, we categorize the state-of-the-art 
into the following three groups, \ie voxel-based, point-based, and point-voxel-based methods. Voxel-based 
methods \cite{zhou2018voxelnet, yan2018second, lang2019pointpillars} generally rasterize irregularly 
distributed point clouds into volumetric grids to extract features from regular domains. Point-based 
methods \cite{shi2019pointrcnn, Yang2020ssd, Point-GNN} directly consume raw point clouds to abstract 
features from irregular domains. Very recently, a large body of investigations have been looking into 
fusing the best of two worlds together for synergies. Point-voxel-based methods \cite{pvrcnn2020, 
he2020sassd, yang2019std} are proposed as a joint treatment for learning more informative representations 
from both regular and irregular domains.
}

\begin{figure}[pos=htbp]
  \centering   
    \includegraphics[height=5.0in,width=3.32in]{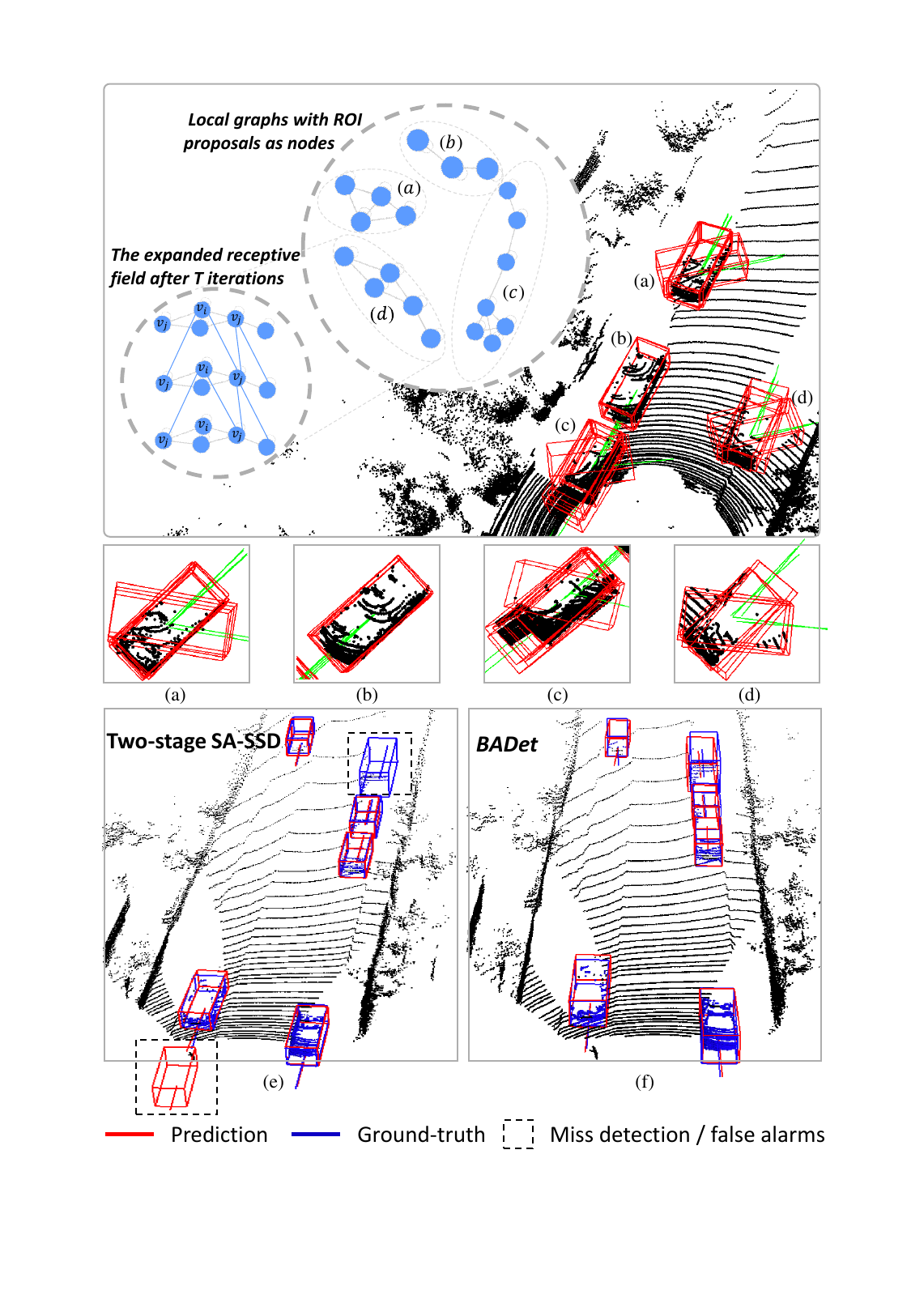} 
    \caption{\highlight{\textbf{ Two-stage SA-SSD \cite{he2020sassd} vs. BADet.} 
    Ideal proposals as shown in (b) are relatively rare. Typical proposals deviate from their 
    ground truth, as shown in (a), (c) and (d).  While proposal refinement is indispensible, the prior art treats 
    the proposals independently without explicitly taking their spatial correlations into account. By contrast, 
    we propose to model such correlations by graph neural networks (GNN), see the top part of the figure. Compared 
    to 3D object detection results of the baseline (e) , our proposed BADet model has less miss detection and fewer 
    false alarms (f). Note that (e) and (f) show different viewpoints of the same point clouds. Two-stage SA-SSD is 
    implemented by directly adding a BEV detect head on the top of the 2D backbone network in the SA-SSD for 
    fair comparison.}
    }
    \label{subfig:whygraph}
    \vspace{-0.2em}
\end{figure}

% \highlight{
Given features extracted from the point clouds, recent high-performance 3D object detectors share a 
two-step working pipeline. First, a region proposal network (RPN) \cite{RenHGS15fasterrcnn} is utilized 
to produce a number of high-quality region-of-interest (RoI) proposals in a bottom-up fashion. RoI 
proposals \wrt a specific object tend to densely appear surrounding the object, as visualized in 
red bounding boxes in Fig. \ref{subfig:whygraph}. Second, RoI pooling is performed separately on each of 
the bounding boxes to obtain RoI-wise representations for further refinement. It is worth pointing out 
that in the previous works, these RoI-wise representations, despite their apparent spatial correlations, 
are treated as uncorrelated entries and thus fed independently to a subsequent detection header. As 
illustrated in Fig. \ref{subfig:whygraph}, proposals in (b) are exactly what we anticipate. 
In contrast, proposals in (a), (c) and (d) all deviate from their ground truth to some extent, 
either in terms of their azimuth or by their positions. Importantly, such deviation becomes even 
more severe at long ranges where point clouds are sparse by nature. Challenges arise in the case where 
a proposal largely forsakes its boundary information due to coordinate offset while existing networks 
lack corresponding information compensation mechanism. As a result, once the deviation occurs, an RoI 
proposal alone is insufficient to capture the boundary information of the underlying object it belongs to. 
On the basis of above discussion and analysis, an important research question arises as \emph{how to exploit 
the spatial correlations among the RoI proposals so that each of them will possess the whole receptive field 
of the associated object?}
% }

% \highlight{
Bearing this in mind, we answer the above question by proposing BADet, a novel \textbf{B}oundary-\textbf{A}ware 
3D object \textbf{Det}ection network. In particular, the object boundary-aware property of our model is achieved 
by graph neural networks (GNN) \cite{Point-GNN} based modeling.  Fig. \ref{subfig:whygraph} shows graph representations 
of the proposals within a specific neighborhood. Different from the existing works, \eg STD \cite{yang2019std}, 
PV-RCNN \cite{pvrcnn2020} and PointRCNN \cite{shi2019pointrcnn}, which refine each proposal independently, 
we consider each proposal in a given neighborhood as a node in a local neighborhood graph. As such, their spatial 
correlations are naturally modeled by GNN feature representation learning. With messages exchanged between the 
nodes in an iterative manner, the receptive field of each proposal is expanded progressively to cover the entire object. 
The GNN learning procedure eventually makes the proposal be aware of the object boundary, even though it initially 
deviates from the ground truth.
% }

\highlight{
Moreover, in order to fully exploit informative semantic features extracted from corresponding regions, we propose 
a lightweight \emph{Region Feature Aggregation} (RFA) module. Specifically, to compensate for the absence of 3D structure 
context when directly converting 3D feature map into BEV representation, we introduce auxiliary branches 
to jointly optimize voxel-wise features from 3D backbone network; to eliminate extensive computation overheads 
when filtering dense proposals as traditional RoIAlign operations do, we interpolate grid point features from  
the nearest pixel along channels for pixel-wise feature extraction \cite{he2020sassd}; to densify gradually 
downsampled 3D feature volumes, we train a PointNet(++) \cite{qi2017pointnet++} network from scratch to inject 
original geometric structure from point-wise features. Our final RoI-wise representations for each node on the 
local neighborhood graph is obtained by aggregating associated voxel-wise, pixel-wise, and point-wise features 
together, before fed through boundary-aware graph neural network for further refinement. 
}
 
Below, we summarize our contributions into three facets. 1) We propose BADet framework which effectively models local 
boundary correlations of an object in the form of local neighborhood graph, which explicitly facilitates a complete 
boundary for each individual proposal by the means of an information compensation mechanism. 2) We propose a lightweight 
region feature aggregation module to make use of informative semantic features, leading to significant improvement 
with manageable memory overheads. 3) We demonstrate the validity of BADet both on widely used KITTI Dataset and 
highly challenging nuScenes Dataset. Our BADet outperforms all previous state-of-the-art methods with remarkable 
margins on KITTI BEV detection leaderboard and ranks $1^{st}$ on $Car$ category of $Moderate$ difficulty as of 
Apr. 17th, 2021. Furthermore, comprehensive experiments are conducted on KITTI Dataset in diverse evaluation 
settings to analyze the effectiveness of BADet.

\vspace{0.5em}
\section{Related Work} \label{sec:related}

\textbf{Representation learning in regular domains.} Voxel-based methods generally voxelize raw point clouds into 
compact volumetric grids, resorting to 3D sparse convolution neural network for representation learning. These 
methods are amenable to hardware implementations while suffer from information loss due to quantization error. 
Limited voxel resolution inevitably hinders more fine-grained localization accuracy. The seminal work 
VoxelNet \cite{zhou2018voxelnet} takes the first lead to rasterize a point cloud into more compact 3D voxel 
volumes and leverages a lightweight PointNet-like \cite{qi2017pointnet,qi2017pointnet++} block to transform points 
within each voxel into a voxel-wise representation, followed by 3D CNNs for spatial context aggregation and 
detections generation. SECOND \cite{yan2018second} utilizes 3D sparse convolution as a substitute of conventional 
3D convolution to only convolve against non-empty voxels. PointPillars \cite{lang2019pointpillars} partitions 
points into ``pillars'' rather than voxels to get rid of the need of convolving against 3D space via forming 
an 2D pseudo BEV image straight forward. 

\textbf{Representation learning in irregular domains.} Point-based methods usually takes raw point clouds 
as inputs. Powered by PointNet \cite{qi2017pointnet,qi2017pointnet++} or Graph Neural Network 
\cite{semigcn2017,graphsage2017}, these methods at utmost preserve 3D structure context from real 3D world 
with flexible receptive fields. Nevertheless, they are hostile to memory overheads
\begin{figure*}[pos=htbp]
	  \includegraphics[width=6.8in,height=3.1in]{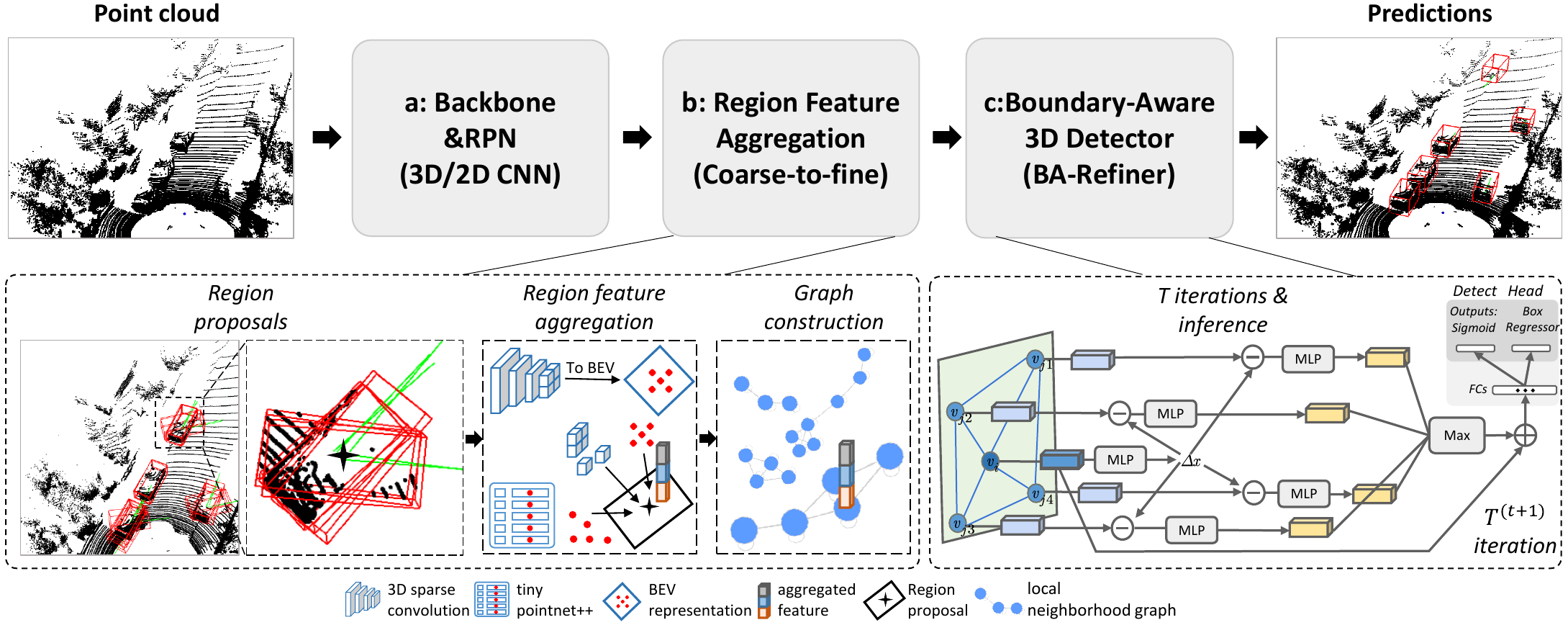}%
	  \caption{\textbf{Overview of our proposed BADet}. BADet includes three key components: (a) Backbone and Region Proposal Network. 
	  We first voxelize the raw point clouds into volumetric grids, resorting to backbone network with a series of 3D 
	  sparse convolutions for multi-scale semantic feature abstraction and 3D proposal generation. 
	  (b) Region Feature Aggregation Module, which fully exploits multi-level semantic features for more informative 
	  RoI-wise representations. 
	  (c) Boundary-Aware Graph Neural Network, which takes 3D proposals in immediate neighborhood
	  as inputs for graph construction within a given cut-off distance, associating 3D proposals 
	  in the form of local neighborhood graph, with boundary correlations of an object being explicitly informed 
	  through an information compensation mechanism. 
	  }
	  \label{subfig:badetoverview}
	  \vspace{-1.5em}
\end{figure*}
and sensitive to translation variance. These methods are typically exemplified by PointRCNN \cite{shi2019pointrcnn}, which 
leverages PointNet(++) \cite{qi2017pointnet++} to segment foreground points from raw point clouds for the 
purpose of reducing 3D search space, elegantly inheriting the ideology of Faster RCNN \cite{RenHGS15fasterrcnn} 
architecture. 3DSSD \cite{Yang2020ssd} safely removes the core feature propagation layer of 
PointNet(++) \cite{qi2017pointnet++} by introducing F-FPS to compensate for the loss of foreground points 
when downsampling. Point-GNN \cite{Point-GNN} seeks to generalize 
graph neural networks (GNNs) to 3D object detection via constructing a graph over downsampled raw point clouds. 
Few investigations have successfully transplanted GNNs for object detection til Shi et al. proposed 
Point-GNN \cite{Point-GNN}. Our BADet differs from Point-GNN by constructing local neighborhood graphs over 
RoI-wise high semantic features rather than raw point clouds. 

\textbf{Representation learning in hybrid domains.} Very recently, a fusion strategy to bring the best of 
two worlds together is increasingly prevalent. Point-voxel-based methods deeply integrate 3D sparse convolution 
operation from voxel-based methods and the flexible receptive fields from point-based methods. PVConv \cite{pvconv2019} 
demonstrates the effectiveness of the combinations of coarse-grained voxel-wise features and fine-grained point-wise 
features for synergies. STD \cite{yang2019std} voxelizes point-wise features for 3D region proposals and 
exploits more generic spherical anchors rather than rectangular ones to achieve a high recall and alleviates 
computational overheads. PV-RCNN \cite{pvrcnn2020} leverages set abstraction operation among voxels instead of 
raw point clouds to achieve flexible receptive fields for fine-grained patterns while maintains computational efficiency.
SA-SSD \cite{he2020sassd} introduces an auxiliary network to consolidate the correlations of 3D feature 
volumes under the supervision of point-level geometric properties. 
\vspace{0.1em}
\section{Proposed BADet} \label{sec:method}

\highlight{
In this section, we propose BADet, which effectively models local boundary correlations of an object by the means 
of local neighborhood graphs, explicitly facilitating a complete boundary for each individual proposal with an 
information compensation mechanism. As illustrated in Fig. \ref{subfig:badetoverview}, the proposed BADet network 
consists of three major components: (a) a backbone followed by a region proposal network (RPN), (b) a region feature 
aggregation (RFA) module and (c) a boundary-aware graph neural network based 3D object detector (BA-Refiner). Specifically, 
we first leverage 3D sparse convolutions to extract features from voxelized point clouds. The features are then 
reshaped to form an BEV representation, which will be fed to the RPN for 3D RoI proposal generation. As for proposal 
refinement, although the BEV representation is informative (see Sec. \ref{EffectofdifferentfeaturecomponentsforRFAmodule}), 
we argue that using it alone suffers from two drawbacks. First, as this representation stems from gradually downsampled 
3D feature volumes, the downsampling operation inevitably dilutes the features. Second, with the depth dimension 
squeezed, the original 3D contextual information is largely lost. To tackle these drawbacks, RFA is introduced to 
aggregate voxel-wise, pixel-wise, and point-wise features, resulting in enriched RoI-wise features. The RoI features 
then go through BA-Refiner for a boundary-aware enhancement. Lastly, the enhanced features are fed to a common 3D 
object detection header to make final detection. 
}

\highlight{
In what follows, we present the backbone and RPN in Sec. \ref{BackboneandRegionProposalNetwork}, followed by RFA 
in Sec. \ref{RegionFeatureAggregationModule} and BA-Refiner in Sec. \ref{BoundaryAwareGraphNeuralNetwork}. Loss 
functions of BADet are detailed in Sec. \ref{LossFunctions}.
}
\vspace{0.5em}
\subsection{Backbone and Region Proposal Network} \label{BackboneandRegionProposalNetwork}
\vspace{0.3em}
\subsubsection{Voxelization} 
\highlight{
We first follow a many-to-one mapping algorithm \cite{he2020sassd,yan2018second,
zhou2018voxelnet} to voxelize raw point clouds for representation learning. Specifically, let $p$ be a point 
in a raw point cloud $\mathcal{P}$ with 3D coordinates $\left( p_x, p_y, p_z \right)$ and reflectance 
intensities $p_r$, where $\mathcal{P}=\left\{ p^i=\left( p_{x}^{i},p_{y}^{i},p_{z}^{i},p_{r}^{i} \right)\! \in \mathbb{R}^4:i=1,...,N \right\}$, 
$N$ indicates the number of points within $\mathcal{P}$. Let $\left[ v_L,v_W,v_H \right] \in \mathbb{R}^3$ be 
the quantization step, we have the voxelized coordinates of $p$ as $(\lfloor \frac{p_{x}}{v_L} \rfloor,\lfloor \frac{p_{y}}{v_W} \rfloor,\lfloor \frac{p_{z}}{v_H} \rfloor)$, 
where $\lfloor \cdot \rfloor $ indicates floor function. A voxel $v$ is formed as a set of points that have 
the same voxelized coordinates. Note that points are stochastically dropped when they exceed the allocated capacity 
for memory cost saving. Accordingly, a point cloud $\mathcal{P}$ is positioned into a feature map with a 
resolution of $L\times W\times H$, subject to the quantization step $\left[ v_L,v_W,v_H \right]$. 
Following \cite{yan2018second, pvrcnn2020}, we obtain the initial feature representation per voxel, 
denoted by $f(v)$, as the mean of its points, \ie $f(v)=\frac{1}{|v|} \sum_{p \in v} (p_x, p_y, p_z, p_r)$.
}
\vspace{0.3em}
\subsubsection{Network architecture} 
\vspace{0.3em}
As shown in Figure \ref{subfig:badetoverview}, we follow the architecture in 
\cite{zhou2018voxelnet,yan2018second,he2020sassd} to employ 3D backbone and 2D 
backbone networks to summarize features for BADet. \highlight{3D backbone network \cite{lang2019pointpillars,pvrcnn2020} 
jointly downscales 3D feature volumes by half from point clouds with a series of stacked 3D sparse convolutions and 
sub-manifold convolutions, each of which follows a batch normalization and non-linear ReLU function.} 2D backbone 
network simply stacks six standard convolutions with kernel size of 3 $\times$ 3, to convolve against 2D bird-view 
feature maps reshaped from the output of 3D backbone network for further feature abstraction. Our detection header 
comprises two sibling 1 $\times$ 1 standard convolutions, generating high-quality 3D proposals in a pixel-by-pixel 
fashion for the following graph construction (see Sec. \ref{NetworkArchitecture}).
\vspace{0.5em}
\subsection{Region Feature Aggregation} \label{RegionFeatureAggregationModule}
The art of feature aggregation for RoI-wise representation is of significance (see Sec. \ref{EffectofdifferentfeaturecomponentsforRFAmodule}). 
As aforementioned, BEV representation is informative. Nevertheless, such an underlying mechanism induces information loss, 
which degrades more accurate localization. Instead, by aggregating voxel-wise, pixel-wise, and point-wise feature 
together, we achieve the top performance. In a sequel, we specify the details of each feature component. 

\subsubsection{Voxel-wise feature encoding} \label{Voxelwisefeatureencoding} 
To obtain voxel-wise feature, we introduce auxiliary branches to jointly optimize voxel-wise features from 3D 
backbone network. \highlight{Specifically, we restore the real-world 3D coordinates for each voxel from nonzero indices 
on the basis of a quantization step at current stage, together with its corresponding 3D sparse convolution 
features, \ie each voxel is in the form of $\left\{ \left( \mathbf{f}_{j}, p_j \right) :j=1,...,M \right\}$, where $\mathbf{f}_{j}$ denotes 
3D sparse convolution feature and $p_j$ indicates the real-world coordinates of a voxel centroid. 
To associate multi-scale features from different stages, we gradually broadcast these 
3D sparse convolution features into raw point clouds via feature propagation algorithm 
mentioned in \cite{qi2017pointnet++}. As the resolution of 3D sparse convolution features decreases, 
background points close to object's boundary are likely to be mistaken for foreground points, and 
consequently mislead the model. To enable our voxel-wise features to be more aware of the object's boundary, 
we add an auxiliary header to the interpolated 3D sparse convolution features of raw point clouds, 
jointly optimizing foreground segmentation and relative center offset estimation by point-wise supervisions. 
After that, we throw the auxiliary header away, and then interpolate 3D sparse convolution 
features $\mathbf{f}^{\left( voxel \right)}$ of raw point clouds from the neck, which compensates for the information 
loss of directly converting 3D sparse convolution feature map into 2D BEV representation without ever restoring 
its 3D structure context.}

\subsubsection{Pixel-wise feature encoding} To obtain our pixel-wise feature, \highlight{we interpolate grid point features 
from the nearest pixel along channels from BEV representation, which is considered as a variant of RoIAlign operation 
adapted from SA-SSD \cite{he2020sassd}. Specifically, for each 3D proposal, we generate $m_1 \times m_2$ 
evenly spaced mesh grids along $x$- and $y$-axes in the BEV perspective of a 3D proposal, respectively. 
Note that $m_1$ and $m_2$ are two user-specified constants which are subject to aspect ratio of a certain category.}
We then encode BEV representation into $m_1\times m_2$ feature maps. For each mesh grid point, we interpolate one 
position over a single feature map via spatial transformer sampler \cite{Transformer} to form the final pixel-wise 
feature component $\mathbf{f}^{\left( pixel \right)}\in \mathbb{R}^{m_1\cdot m_2}$ of RoI-wise representation 
at negligible cost. Pixel-wise feature can be extracted by any traditional RoIAlign operations 
in practice. Whereas, traditional pooling mechanism suffers from expensive computation overheads.  

\subsubsection{Point-wise feature encoding} To compensate for the information loss induced by quantization 
error \cite{pvrcnn2020, Yang2020ssd}, we employ an adapted PointNet(++) \cite{qi2017pointnet++} variant. 
As we leverage the commonly used backbone \cite{zhou2018voxelnet, yan2018second, lang2019pointpillars} 
to gradually 8 times downsample 3D feature volumes for the purpose of memory usage saving, which dilutes 
3D features inevitably. We therefore train a tiny PointNet from scratch, which consumes raw point clouds 
as inputs to summarize the whole scene into a fraction of keypoints' semantic features, which are then broadcast 
into the centroids of the proposed 3D proposals to obtain point-wise feature component, 
$\mathbf{f}^{\left( point \right)}$, with the help of feature propagation algorithm in \cite{qi2017pointnet++}. 

Finally, we summarize the three aforementioned multi-level associated feature components via concatenation to 
enrich the final RoI-wise representations for graph construction in the stage two, which significantly 
contributes to the performance (see Sec. \ref{Effectofboundaryaware3DdetectorwithTiterations}) as 
\begin{equation}
	\begin{aligned}
		\text{X}\!=\!\left\{\mathbf{x}_{i}\!=\!\left(x_i,\!\left[ \mathbf{f}_{i}^{\left( voxel \right)},\mathbf{f}_{i}^{\left( pixel \right)},\mathbf{f}_{i}^{\left( point \right)} \right] \right)\!\!:i\!=\!1,...,n \right\}. 
	   \label{eq:roiwisefeature}
	\end{aligned}      
\end{equation}
\vspace{0.5em}
\subsection{Boundary-Aware 3D Detector} \label{BoundaryAwareGraphNeuralNetwork}
In this section, we describe our boundary-aware 3D detector. We assume that we have already obtained 
3D proposals and their corresponding RoI-wise representations. We first describe how the graph is 
constructed (see Sec. \ref{Graphconstruction}). We then detail graph update algorithm (see Sec. \ref{Graphupdate}).

\subsubsection{Graph construction} \label{Graphconstruction} 
\highlight{
Consider an $(F+3)$-dimensional detection set with $n$ 3D proposals, denoted by 
$\text{X}=\left\{ \mathbf{x}_1,...,\mathbf{x}_{\mathbf{n}} \right\} \subseteq \mathbb{R}^{F+3}$, 
where $\mathbf{x}_i=(x_i, \mathbf{s}_i)$ is regarded as a node of graph $\mathcal{G}$ with 
3D coordinates $x_i\in \mathbb{R}^3$ and initial node state $\mathbf{s}_i \in \mathbb{R}^F$. 
Note that coordinates $x_i$ represents the centroid of a detected 3D proposal and 
the node state $\mathbf{s}_i$ is initialized accordingly by the RoI-wise representation extracted and 
aggregated from the corresponding region. }In our case, we construct local neighborhood 
graph $\mathcal{G}\left( \mathcal{V},\mathcal{E} \right) $ of $\text{X}$ in $\mathbb{R}^{F+3}$ as
\begin{equation}
	\begin{aligned}
		\mathcal{E}=\left\{ (i, j)|\lVert x_i-x_j \rVert _2<r \right\}, 
		\label{eq:graph}
	\end{aligned}      
\end{equation}
where $\mathcal{V}=\left\{ 1,...,n \right\} $ and $\mathcal{E}\subseteq \mathcal{V}\times \mathcal{V}$ are the nodes and edges, respectively, $r$ is the given cut-off threshold. 
$\mathcal{G}\left( \mathcal{V},\mathcal{E} \right) $ is undirected, which includes self-loop, meaning each node connects to itself. In this work, 
we define $\mathcal{N}\left( i \right) $ as neighborhood function, which draws neighbors from the set 
$\left\{ j\in \mathcal{V}\ :\ \left( i,\ j \right) \in \mathcal{E} \right\}$ 
with a runtime complexity $O\left( c\left| \mathcal{V} \right| \right) $ in the worst case, where $c$ is 
an user-specified constant.

\subsubsection{Graph update} \label{Graphupdate}
\highlight{
\textbf{The vanilla graph update algorithm.} 
The main idea described in Algorithm \ref{Graphupdatealgorithm} is that each node of 
$\mathcal{G}(\mathcal{V}, \mathcal{E} )$ updates itself via aggregating information from 
their immediate neighbors at the previous iteration. As information flows over nodes, 
each node gradually gains an expanding receptive field from further reaches of its neighbors. 
As we have illustrated in Sec. \ref{sec:intro}, 3D proposals $\text{X}$ that generated by region proposal network 
usually offset from ground truth somehow, emerging in local neighborhood densely with an underlying probability. 
In fact, only a handful of proposals should fully cover the regions where objects truly exist.  
\begin{algorithm} 
	\caption{The vanilla graph update algorithm} 
	\label{Graphupdatealgorithm}
	\KwIn{Graph $\mathcal{G}(\mathcal{V}, \mathcal{E} )$; 
		  input features $\{\mathbf{x}_i , \forall i \in \mathcal{V} \}$; depth $K$; 
		     weight matrices $\mathbf{W}^k_{g},\mathbf{W}^k_{f}$, $\forall k \in \{1, \cdots, K\}$; 
			 non-linearity $\sigma$; channel-wise symmetric aggregator function ${\Box}_k \left( \text{\eg}\ \Sigma \ \text{or\ }\max \right), \forall k \in \{1, \cdots, K\}$; 
			 concatenation function: $CAT$ along aixs $1$ dimension;
			 neighborhood function $\mathcal{N} : i \to 2^\mathcal{V}$.}
	\KwOut{Vector representations $\mathbf{z}_i$ for all $i \in \mathcal{V}$.}
	$\mathbf{h}_i^0\gets \mathbf{x}_i, \forall i\in \mathcal{V}$\; 
	\For{$k=1\dots K$} 
	{ 
		\For{$i\in \mathcal{V}$} 
		{ 
			$\mathbf{h}_{\mathcal{N}\left( i \right)}^{k}\gets\! \Box _k\!\left( \mathbf{W}^k_{g}\!\cdot\! CAT\!\left(\! \{\mathbf{h}_{j}^{k-1},\forall j\in \mathcal{N}\left( i \right) \}\! \right)\! \right)$\;
			$\mathbf{h}_{i}^{k}\gets \mathbf{h}_{i}^{k-1}+\sigma \left( \mathbf{W}^k_{f}\cdot \mathbf{h}_{\mathcal{N}\left( i \right)}^{k} \right) $\;
		} 
	} 
	$\mathbf{z}_i\gets \mathbf{h}_i^K, \forall i\in \mathcal{V}$. 
\end{algorithm}  
More often than not, proposals $\text{X}$ we obtained can deviate somehow in cases where azimuth or $x_i$ offsets 
from the ground truth. As a result, when we extract and aggregate semantic features from these proposed regions 
to summarize RoI-wise representations for refinement, a proportion of the boundary information of 
an object is naturally lost. Under existing coarse-to-fine paradigms, RoI-wise representations are treated 
individually as uncorrelated entries when fed to following detection headers, each of which alone fails to  
perceive a complete boundary of an object by itself. In our case, we model local boundary 
correlations of an object in the form of local neighborhood graph to exploit the spatial correlations among the 
RoI proposals so that each of them will possess the whole receptive field of the associated object, \ie 
``boundary aware", by taking the semantic features from further reaches of its neighbors into account.}
As described in Algorithm \ref{Graphupdatealgorithm}, $k$ in the outer loop represents the current step, 
$\mathbf{h}_k$ indicates the hidden state of a node at present step. In the vanilla version, 
$\left\{ \mathbf{h}_{j}^{k-1},\forall j\in \mathcal{N}\left( i \right) \right\}$ denotes associated feature
vectors of the local neighbors of node $i \in \mathcal{V}$, which is directly concatenated before fed through a 
multi-layer perceptron network (MLP) for feature transformation. We then aggregate these transformed 
features into a single vector, $\mathbf{h}_{\mathcal{N}\left( i \right)}^{k}$, via aggregation operator. 
In this paper, we adopt $\max \left( \cdot \right) $ aggregator inspired by rencent advances in 
leveraging graph neural network to learn feature representations over point sets \cite{Point-GNN}. 

\highlight{
\textbf{The extended graph update algorithm.} 
To alleviate translation variance mentioned in \cite{Yang2020ssd,Point-GNN}, we extend our graph propagation 
strategy via concatenating the relative offset $x_i-x_j$ to the semantic feature of each node $j \in \mathcal{V}$. 
Note that node $j$ is from further reaches of node $i$'s neighbors, where $i \in \mathcal{V}$ is the central node. 
Also, given $\mathbf{x}_i$ already contains local structure context from previous iteration, we follow \cite{Point-GNN} 
to predict an alignment offset $\varDelta x_{i}^{k}$ for each node in $\mathcal{N}\left( i \right)$. These 
alignment offsets are insensitive to slight disturbances or jitteriness, and thus lead to more robust detection outcomes.
Let $\left[ \cdot \right] $ be 
an element-wise concatenation and thus the final aggregated feature in this paper is obtained by as
\begin{equation}
	\begin{aligned}
		\mathbf{h}_{\mathcal{N}\left( i \right)}^{k}\!\!\!\gets \!\max_k\!\left(\! \mathbf{W}_{g}^{k}\!\cdot\!CAT\!\left(\! \left\{\! \left[ x_i\!-\!x_j\!-\!\varDelta x_{i}^{k};\mathbf{h}_{j}^{k-1} \right] \!,\!\forall j\!\in \!\mathcal{N}\left( i \right)\! \right\}\! \right)\! \right). \;
		\label{eq:1}
	\end{aligned}      
\end{equation}
}
\vspace{-2em}
\subsubsection{Graph header} We introduce two sibling branches, which 
takes $\mathbf{z}_i$ as inputs with two stacked MLP layers for refinement:
one is for bounding box classification, and the other is for more accurate oriented 3D bounding box regression 
(see Figure \ref{subfig:badetoverview}).

\subsection{Loss Functions} \label{LossFunctions}
To learn informative representations in a fully supervised fashion, we follow the conventional anchor-based 
settings in \cite{yan2018second}. In particular, we use Focal Loss \cite{LinGGHD17RetinaNet}, Smooth-L1 Loss 
for the bounding box classification and regression in both stage one ($\mathcal{L}_{rpn}$) and 
stage two ($\mathcal{L}_{gnn}$), respectively. \highlight{To obtain better boundary-aware voxel-wise representations described in 
Sec. \ref{Voxelwisefeatureencoding}, we introduce a center offset estimation branch as
\begin{equation}
	\begin{aligned}
		\mathcal{L}_{offset}&\!=\!\frac{1}{N_{pos}}\sum_i^N{\mathcal{L}_{\text{smooth}-\text{L}1}\!\left(\! \varDelta \hat{d}\!-\!\varDelta d^*\! \right)\! \cdot\! \mathbbm{1}\left[ b_{i}^{*}\ge 1 \right]},
		\label{eq:}
	\end{aligned}      
\end{equation}
where $\varDelta d^*$ indicates the offsets of interior points of a ground-truth bounding box from center 
coordinates, $N_{pos}$ represents the number of foreground points, $\mathbbm{1}\left[ b_{i}^{*}\ge 1 \right]$ indicates 
only the points fall in ground-truth bounding box are considered. Besides, we summarize foreground segmentation loss as
\begin{equation}
	\begin{aligned}
		\mathcal{L}_{seg}&=\frac{1}{N_{pos}}\sum_i^N{-}\alpha \left( 1-\hat{p}_i \right) ^{\gamma}\log \left( \hat{p}_i \right),  \\
		\label{eq:1}
	\end{aligned}
	\vspace{-1em}      
\end{equation}
where $\hat{p}_i $ indicates the predicted foreground probability. }The overall loss $\mathcal{L}$ is formulated as
\begin{equation}
	\begin{aligned}
		\mathcal{L}=\mathcal{L}_{rpn}+\mathcal{L}_{gnn}+\mathcal{L}_{offset}+\mathcal{L}_{seg}.
		\label{eq:1}
	\end{aligned}      
\end{equation}
\vspace{0.5em}                    
\section{Evaluation} \label{sec:eval}

In this section, we introduce experimental setup of BADet, including 
dataset, network architecture, as well as training and inference details (Sec. \ref{ExperimentalSetup}). 
\highlight{We report comparisons with the existing state-of-the-art approaches on both the 
widely used KITTI Dataset (Sec. \ref{ResultsontheKITTIDataset}) and highly challenging nuScenes 
Dataset \ref{ResultsonthenuScenesDataset}. }In Sec. \ref{AblationStudies}, extensive ablation studies are 
conducted on KITTI Dataset in diverse evaluation settings to investigate the effectiveness of each 
component of BADet. Also, we report runtime analysis for future optimization (Sec. \ref{RuntimeAnalysis}).

\subsection{Experimental Setup} \label{ExperimentalSetup}
\vspace{0.5em}
\subsubsection{Dataset} \textbf{KITTI Dataset} \cite{geiger2013vision,geiger2012we} contains 7,481 training 
samples and 7,518 testing samples for three categories (\ie $Car$, $Pedestrian$, and $Cyclist$) in the 
context of autonomous driving, each of which has three difficulty levels (\ie $Easy$, $Moderate$, and $Hard$) 
according to object scale, occlusion, and truncation levels. As a common practice mentioned in \cite{chen2017multi}, 
we evaluate our proposed BADet on the KITTI 3D and BEV object detection benchmark following the frequently 
used \emph{train}, \emph{val} split, with 3,712, 3,769 annotated samples for training, validation, respectively. 
Following previous literature \cite{he2020sassd, ciassd}, we report the performance 
of the most commonly-used car category for fair comparison on both the \emph{val} split and 
the \emph{test} split by submitting to the online learderboard.

\highlight{
\textbf{nuScenes Dataset} \cite{nuscenes2019} consists of 1000 challenging driving sequences, each of which 
is about 20-second long, with 30k points per frame. It totally annotates 1.17M 3D objects on 10 different 
classes with 700, 150, 150 annotated sequences for \emph{train}, \emph{val} and \emph{test} split, 
respectively. In particular, 28k key frames are annotated for \emph{train} split, and 6k, 6k for \emph{val} 
and \emph{test} split accordingly. To predict velocity, existing methods concatenate lidar points from key
frame and frames in last 0.5-second together, resulting in approximately 400k points in total.  
Following previous literature \cite{Yang2020ssd}, we train BADet on \emph{train} split and compare with the 
state-of-the-art methods on \emph{val} split to further confirm the validity of the proposed method.
}

\subsubsection{Network Architecture} \label{NetworkArchitecture} As shown in Figure \ref{subfig:visualiztion_part1}, 
we follow the architecture in \cite{zhou2018voxelnet,yan2018second,he2020sassd} to design 3D backbone and 2D 
backbone network for BADet. 3D backbone network downsamples 3D feature volumes from point clouds with 
dimensions 16, 32, 64, 64, respectively, each of which comprises a series of 3 $\times$ 3 $\times$ 3 sub-manifold 
sparse convolutions, together with a batch normalization and non-linear ReLU following each convolution. 
2D backbone network stacks six standard 3 $\times$ 3 convolutions, of which filter number is 256, to 
convolve against 2D bird-view feature map reshaped from the output of 3D backbone 
network for further feature abstraction. The detection header is composed of two sibling 1 $\times$ 1 
convolutions with filter numbers of 256, which generates high-quality 3D proposals for
graph construction in the second stage. As for region feature aggregation module, we use MLPs of 
unit (32+64+64, 64) for feature transformation, followed by two branches of units (64, 1), (64, 3) for 
foreground classification and center offset estimation. Also, we train a lightweight tiny PointNet-like block 
from scratch, of which two neighboring radii of each level are set as (0.1m, 0.5m), (0.5m, 1.0m), (1.0m, 2.0m), respectively. 
The keypoints sampled for each level is set to 4,096, 1,024, 256 via FPS algorithm \cite{qi2017pointnet++}. 
Finally, with regard to boundary-aware 3D detector, we stack three MLPs of units 
(64+28+28, 60, 30, 120) with ReLU non-linearity for computating edges and updating nodes of local neighborhood graphs. 

\subsubsection{Training} Our BADet consumes regular voxels as input. For KITTI Dataset, we first voxelize the 
raw point clouds with a quantization step of [0.05m, 0.05m, 0.1m]. Note that only objects in FOV are annotated for 
KITTI Dataset, we follow \cite{lang2019pointpillars,ciassd,he2020sassd} to clip the range of point clouds 
into [0, 70.4]m, [-40, 40]m, and [-3, 1]m along the x, y, z axes, respectively. The resolution of final BEV 
feature map is 200 $\times$ 176. Hence, a total of 200 $\times$ 176 $\times$ 2 pre-defined anchors 
(width=1.6m, length=3.9m, height=1.56m) for car category are evenly generated, with two possible orientations 
($0^\circ$ or $90^\circ$) being considered. These anchors are subject to the averaged dimensions over the whole 
KITTI Dataset. Furthermore, we follow the matching strategy of VoxelNet\cite{zhou2018voxelnet} to distinguish 
the negative and positive anchors with IoU thresholds 0.45 and 0.6, respectively. \highlight{For nuScenes Dataset, the 
detection range is within [-54, 54]m for the x axis, [-54, 54]m for the y axis and [-5, 3]m for z axis, with 
a quantization step of [0.075m, 0.075m, 0.2m] along each axis. Given that nuScenes Dataset has 10 classes 
with a large variance on scale, we therefore use multiple seperated regression heads for 
region proposals generation in the first stage.}

The whole architecture of our BADet is optimized from scratch in an end-to-end fashion. 
For KITTI Dataset, we optimize the entire network with batch size 2, weight decay 0.001 for 
50 epochs with the SGD optimizer on a single GTX 1080 Ti GPU. The learning rate is initialized with 0.01, 
which is decayed with a cosine annealing strategy adopted in SA-SSD \cite{he2020sassd}. During training, 
we consider frequently adopted data augmentations to facilitate our BADet's generalization ability, including 
global rotation around $z$-axis with the noise uniformly drawn from $\left[ -\frac{\pi}{4},\frac{\pi}{4} \right]$, 
global scaling with 
\begin{table*}[htbp]
	\centering
	\caption{\textbf{Performance comparison with state-of-the-arts on KITTI test server.}
  We report the Average Precision(AP) with 40 recall positions on both BEV and 3D object detection 
  leaderboard of $Car$ category with a rotated IoU threshold 0.7. The top performance is indicated in bold.
  Please refer to 
  \href{http://www.cvlibs.net/datasets/kitti/eval_object_detail.php?&result=48db930a7077e9925311b2539c21aed7541b7295} 
	{\emph{\text{http://www.cvlibs.net/kitti/eval\_object?benchmark=3d}}} for the purpose of 
  online validation.
  }
%   \vspace{-0.8em}
 % \vspace{0.7em}
	\label{tab:test}%
	  \begin{tabular}{lcccccc}
	  \toprule
	  \multicolumn{1}{c}{\multirow{2}[2]{*}{\textbf{Method}}} & \multicolumn{3}{c}{$AP_{3D}\left( \% \right) $} & \multicolumn{3}{c}{$AP_{BEV}\left( \% \right) $} \\
          & \multicolumn{1}{c}{\textbf{Easy} $\downarrow$} & \multicolumn{1}{c}{\textbf{Mod.} $\downarrow$} & \multicolumn{1}{c}{\textbf{Hard} $\downarrow$} & \multicolumn{1}{c}{\textbf{Easy} $\uparrow$} & \multicolumn{1}{c}{\textbf{Mod.} $\uparrow$} & \multicolumn{1}{c}{\textbf{Hard} $\uparrow$} \\
	  \midrule
	  VoxelNet\cite{zhou2018voxelnet}, CVPR18 & 77.82 & 64.17 & 57.51 & 87.95 & 78.39 & 71.29 \\
	  SECOND\cite{yan2018second}, Sensors18 & 83.34 & 72.55 & 65.82 & 89.39 & 83.77 & 78.59 \\
	  PointPillars\cite{lang2019pointpillars}, CVPR19 & 82.58 & 74.31 & 68.99 & 90.07 & 86.56 & 82.81 \\
	  Point-GNN\cite{Point-GNN}, CVPR20 & 88.33 & 79.47 &72.29 & 93.11 & 89.17 & 83.90 \\
	  3DSSD\cite{Yang2020ssd}, CVPR20 & 88.36 & 79.57 & 74.55 & 92.66 & 89.02 & 85.86 \\
	  SA-SSD\cite{he2020sassd}, CVPR20 & 88.75 & 79.79 & 74.16 & 95.03 & 91.03 & 85.96 \\
	  \midrule
	  MV3D\cite{chen2017multi}, CVPR17 & 74.97 & 63.63 & 54.00    & 86.62 & 78.93 & 69.80 \\
	  ContFuse\cite{ContFuse2018ming}, ECCV18 & 83.68 & 68.78 & 61.67 & 94.07 & 85.35 & 75.88 \\
	  F-PointNet\cite{qi2018frustum}, CVPR18 & 82.19 & 69.79 & 60.59 & 91.17 & 84.67 & 74.77 \\
	  AVOD-FPN\cite{ku2018avod}, IROS18 & 83.07 & 71.76 & 65.73 & 90.99 & 84.82 & 79.62 \\
	  PointRCNN\cite{shi2019pointrcnn}, CVPR19 & 86.96 & 75.64 & 70.70  & 92.13 & 87.39 & 82.72 \\
	  MMF\cite{mmf2019cvpr}, CVPR19 & 88.40  & 77.43 & 70.22 & 93.67 & 88.21 & 81.99 \\
	  3D-CVF\cite{3dcvf2020jin}, ECCV20 & 89.20  & 80.05 & 73.11 & 93.53 & 89.56 & 82.45 \\
	  3D IoU Loss\cite{iouloss3d}, 3DV19 & 86.16 & 76.50  & 71.39 & 91.36 & 86.22 & 81.20 \\
	%   Fast PointRCNN\cite{Chen0SJ19}, ICCV19 & 85.29 & 77.40  & 70.24 & 90.87 & 87.84 & 80.52 \\
	  Part-A$^2$\cite{shi2020points2parts}, TPAMI20 & 87.81 & 78.49 & 73.51 & 91.70  & 87.79 & 84.61 \\
	  STD\cite{yang2019std}, ICCV19 & 87.95 & 79.71 & 75.09 & 94.74 & 89.19 & 86.42 \\
	  PV-RCNN\cite{pvrcnn2020}, CVPR20 & 90.25 & 81.43 & 76.82 & 94.98 & 90.65 & 86.14 \\
	  Voxel R-CNN\cite{voxelrcnn}, AAAI21 & \textbf{90.90} & \textbf{81.62} & \textbf{77.06} & 94.85 & 88.83 & 86.13 \\
	%   \midrule
	%   \midrule
	  \emph{BADet} & 89.28 & 81.61 & 76.58 & \textbf{95.23} & \textbf{91.32} & \textbf{86.48} \\
	%   Improvement $\left( \Uparrow \% \right) $ & \cellcolor[rgb]{ .851,  .851,  .851} $-1.62$ & \cellcolor[rgb]{ .851,  .851,  .851} $-0.01$ & \cellcolor[rgb]{ .851,  .851,  .851} $-0.48$ & \cellcolor[rgb]{ .851,  .851,  .851} $+0.20$ & \cellcolor[rgb]{ .851,  .851,  .851} $+0.29$ & \cellcolor[rgb]{ .851,  .851,  .851} $+0.52$ \\
	  \bottomrule
	  \end{tabular}%}%
	%   \vspace{-1.9em}
\end{table*}%
\begin{table}[htbp]
	\centering
	\caption{\textbf{Performance comparison with state-of-the-arts on val set.}
	  We report the Average Precision(AP) with 11 recall positions on 3D object detection 
	  leaderboard of $Car$ category with a rotated IoU threshold 0.7. The top performance is indicated in bold.}
	\label{tab:val11}%
	% \vspace{-0.8em}
	% \vspace{-0.7em}
	\setlength{\tabcolsep}{0.7mm}{
	\begin{tabular}{lccc}
	\toprule
	\multicolumn{1}{c}{\multirow{2}[2]{*}{\textbf{Method}}} & \multicolumn{3}{c}{$AP_{3D}\left( \% \right) $} \\
		& \multicolumn{1}{c}{\textbf{Easy} $\downarrow$} & \multicolumn{1}{c}{\textbf{Moderate} $\uparrow$} & \multicolumn{1}{c}{\textbf{Hard} $\downarrow$} \\
	\midrule
	SECOND\cite{yan2018second}, Sensors18 & 87.43 & 76.48 & 69.10 \\
	PointPillars\cite{lang2019pointpillars}, CVPR19 &   -    & 77.98 & - \\
	PointRCNN\cite{shi2019pointrcnn}, CVPR19 & 88.88 & 78.63 & 77.38 \\
	3DSSD\cite{Yang2020ssd}, CVPR20 & 89.71 & 79.45 & 78.67 \\
	CIA-SSD\cite{ciassd}, AAAI21 & 90.04 & 79.81 & 78.80 \\
	SA-SSD\cite{he2020sassd}, CVPR20 & \textbf{90.15} & 79.91 & 78.78 \\
	  \midrule
	  MV3D\cite{chen2017multi}, CVPR17 & 71.29 & 62.68 & 56.56 \\
	  F-PointNet\cite{qi2018frustum}, CVPR18 & 83.76 & 70.92 & 63.65 \\
	  ContFuse\cite{ContFuse2018ming}, ECCV18 & 86.32 & 73.25 & 67.81 \\
	  AVOD-FPN\cite{ku2018avod}, IROS18 & 84.41 & 74.44 & 68.65 \\
	  Fast Point R-CNN\cite{Chen0SJ19}, ICCV19 & 89.12 & 79.00 & 77.48 \\
	  STD\cite{yang2019std}, ICCV19 & 89.70 & 79.80 & \textbf{79.30} \\
	  PV-RCNN\cite{pvrcnn2020}, CVPR20 &    -   & 83.90 & - \\
	  Voxel R-CNN\cite{voxelrcnn}, AAAI21&89.41 & 84.52 & 78.93 \\
	%   \midrule
	%   \midrule
	  BADet & 90.06 & \textbf{85.77} & 79.00 \\
	  \bottomrule
	  \end{tabular}}%
	%   \vspace{-1.5em}
  \end{table}% 
the scaling factor uniformly drawn from [0.95, 1.05], and global flipping along $x$-axis 
simultaneously. Also, we adopt a grond-truth augmentation strategy proposed by SECOND \cite{yan2018second} to 
enrich the current training scene by seamlessly ``pasting'' a handful of new ground-truth boxes and the 
associated points that fall in them. \highlight{For the nuScenes Dataset, we optimize the entire network with batch size 2, 
weight decay 0.01 for 7 epochs with the ADAM optimizer on two GTX 1080 Ti GPUs. We adopt the one cycle learning rate 
strategy with default settings, where the maximum learning rate is 0.001. }

\subsubsection{Inference} For KITTI Dataset, we select high-confidence bounding boxes with a threshold 0.3. We use 
non-maximum suppression (NMS) with a rotated IoU threshold 0.1 to remove redundancy. Please refer to 
BADet's source code for more details since we will enclose herewith implementation details. 

\vspace{0.5em}
\subsection{3D Detection on the KITTI Dataset} \label{ResultsontheKITTIDataset}
To evaluate the performance of BADet on \emph{val} set, 
we utilize the \emph{train} split for training. To fairly evaluate the proposed BADet's performance 
on \emph{test} set, of which the labels are unavailable, we train on all \emph{train+val} data and 
use the parameters of last epoch for online test server submission. 

\subsubsection{Evaluation Metric} We follow \emph{KITTI Dataset} protocal by using 
\emph{Interpolated AP@$S_N$ Metric}, which indicates the mean precision of N 
equally spaced recall levels as
\begin{equation}
  \begin{aligned}        
      \text{AP}=\frac{1}{N}\sum_{r\in S}{P_{interpolate}\left( r \right)},
      \label{eq:1}
  \end{aligned}      
\end{equation} 
where $S=\left[ q_0,q_0+\frac{q_1-q_0}{N-1},q_0+\frac{2\left( q_1-q_0 \right)}{N-1},...,q_1 \right]$ is a 
subset for N recall levels, of which the precision interpolated for each recall level $r$ is represented by 
\begin{equation}
  \begin{aligned}        
      P_{interpolate}\left( r \right) =\underset{\tilde{r}:\tilde{r}\ge r}{\max}\ P\left( \tilde{r} \right).
      \label{eq:1}
  \end{aligned}      
\end{equation} 
\highlight{
As a common practice, we report average precision (AP) with a rotated IoU threshold 0.7 for $Car$ in terms 
of assessing the quality of our proposed BADet. From 08.10.2019, KITTI benchmark adopts 
the \emph{Interpolated AP@$S_{40}$ Metric} with 40 recall levels $S_{40}=\left[ 1/40,2/40,3/40,...,1 \right]$ as 
suggested in \cite{SimonelliBPLK19} for more fair evaluation. We follow the conventions as previous works 
\cite{he2020sassd,pvrcnn2020,ciassd} do. We evaluate on the $test$ set by submitting to online $test$ server 
of which 40 recall positions are considered. We evaluate on $val$ set with 11 recall positions 
$S_{11}=\left[ 0,0.1,0.2,...,1 \right]$ to compare with existing state-of-the-arts unless otherwise noted.}
  
\subsubsection{Comparison with State-of-the-Arts}
\begin{figure*}[pos=htbp]
	\centering   
	  \includegraphics[width=5in]{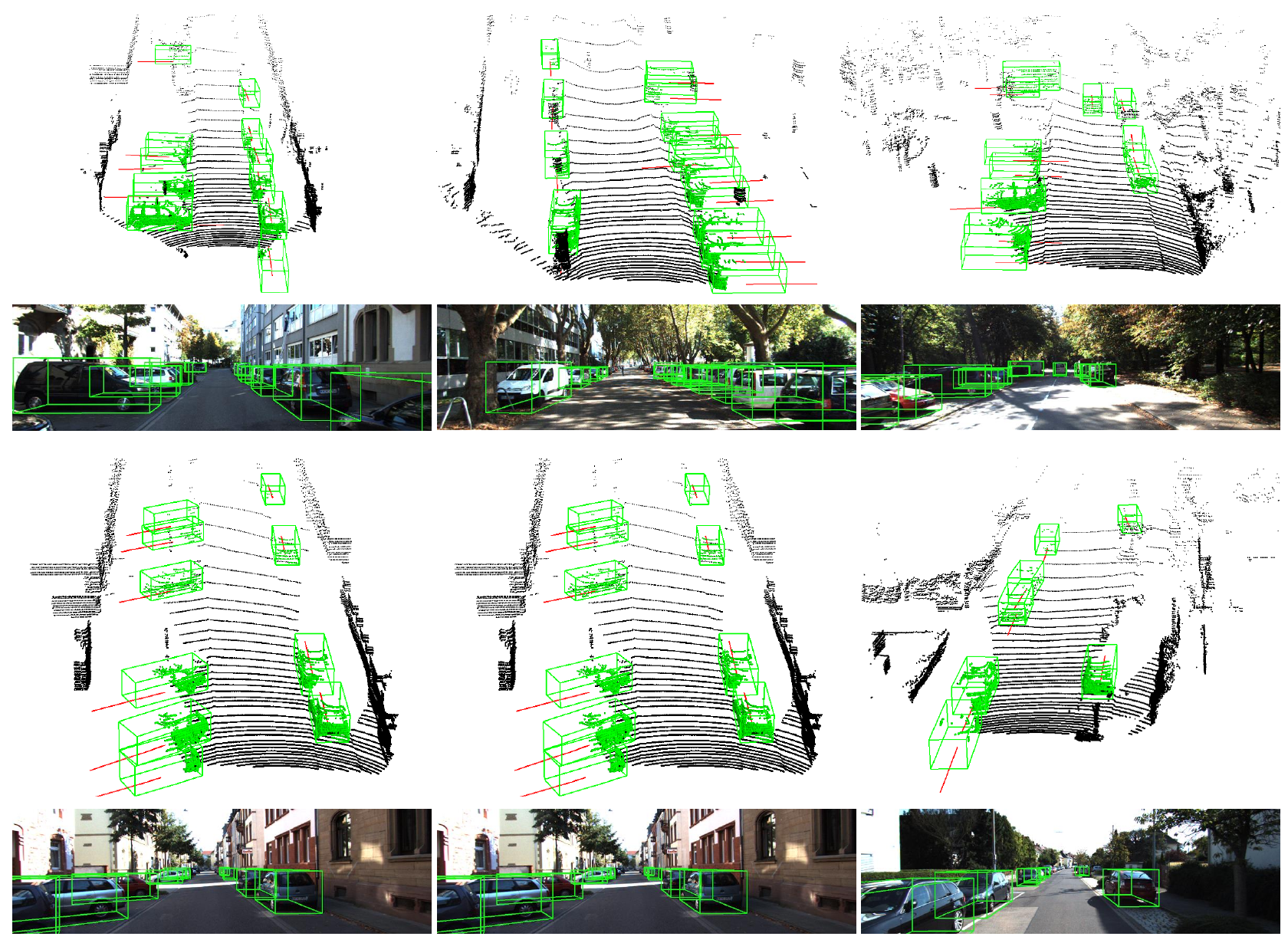} 
	%   \vspace{-0.8em}
	  \caption{\textbf{Qualitative results on KITTI test set using BADet.} The predicted bounding boxes of Cars (in 
	  green) are drawn on both the point cloud ($1^{st}$ \& $3^{rd}$ ) and the image ($2^{nd}$ \& $4^{th}$). 
	  Best view in color.}
	  \label{subfig:visualiztion_part1}
	  \vspace{-1.5em}
\end{figure*}

\begin{figure*}[pos=htbp]
    \centering  
    % \subfigure[paper overview]{ 
      \includegraphics[width=0.7\linewidth]{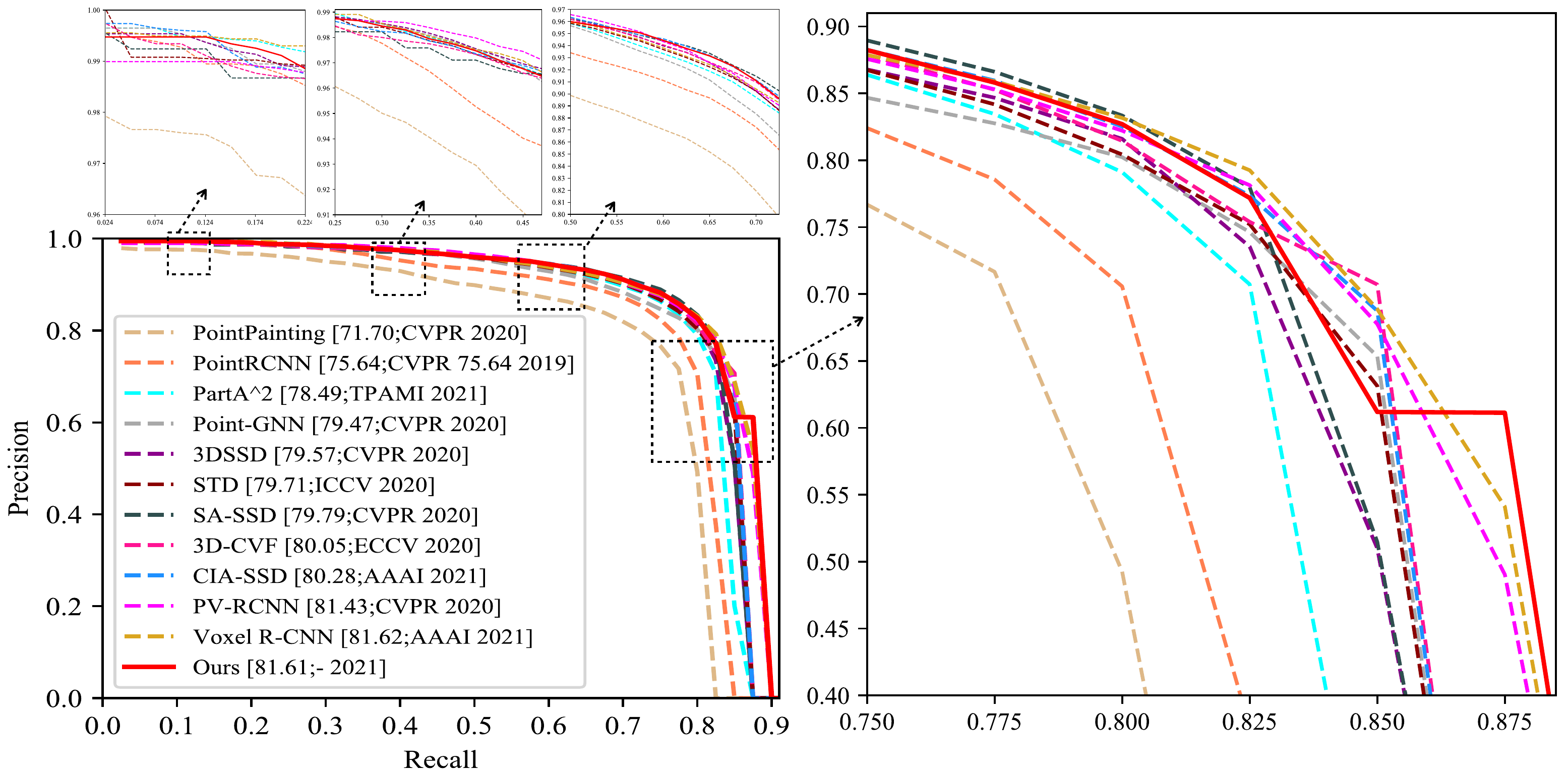} 
	%   \vspace{-0.8em}
      \caption{\textbf{Precision $\times$ Recall curve of state-of-the-art methods on online KITTI 
	  3D detection benchmark.} We report average precision (AP) with a rotated IoU threshold 0.7 for \emph{Moderate} 
	  difficulty of \emph{Car} category.}
      \label{subfig:prcurve} %% label for first subfigure 
	  \vspace{-1em}
    % }    
\end{figure*}

We compare the performance of our BADet on the KITTI \emph{test} set with previous state-of-the-art methods 
by submitting predictions to the official online leaderboard \cite{geiger2012we,geiger2013vision}. As shown in 
Table \ref{tab:test}, Our BADet outperforms all its competitors with remarkable margins on KITTI BEV detection 
leaderboard and ranks $1^{st}$ on $Car$ category of $Moderate$ difficulty, surpassing the second place, SA-SSD 
\cite{he2020sassd} by $\left(0.2 \%, 0.29 \%, 0.52 \% \right)$ for \emph{Easy}, \emph{Moderate}, and \emph{Hard} 
level respectively as of Apr. 17th, 2021. Note that our backbone network at the first stage is built upon the 
architecture of SA-SSD while obtains an absolute gain by (0.53\%, 1.82\%, 2.42\%) on 3D detection benchmark. 
To the best of our knowledge, little literature 
has achieved decent performance on 3D object detection via leveraging graph neural network until 
Point-GNN \cite{Point-GNN} is proposed. We surpass Point-GNN (0.95\%, 2.14\%, 4.29\%) on 3D detection 
benchmark, (2.12\%, 2.15\%, 2.58\%) on BEV detection benchmark while keep nearly 5 times faster than Point-GNN 
(7.1 FPS vs. 1.5 FPS). Compared with the latest state-of-the-art Voxel R-CNN \cite{voxelrcnn}, 
our BADet is slightly lower (-0.01\%) but remains close on 3D detection benchmark while achieves 
(0.38\%, 2.49\%,, 0.35\%) improvement on BEV detection benchmark. Note that we outperform Voxel R-CNN 
by 2.49\% on \emph{Car} category of \emph{Moderate} difficulty. Figure \ref{subfig:prcurve} further indicates 
better detection coverage of our  BADet with different recall settings. Furthermore, we visualize some 
predictions from \emph{test} set for qualitative comparison in Figure \ref{subfig:visualiztion_part1}.  

% \textbf{KITTI-val.} 
In addition, as shown in Figure \ref{tab:val11}, we exploit \emph{Interpolated AP@$S_{11}$ Metric} with a 
rotated IoU threshold 0.7 for further comparison with previous works on \emph{val} set. Besides, by considering 
the results in Table \ref{tab:test}, \ref{tab:val11}, we argue that such slightly inconsistent result between  
\emph{val} set and \emph{test} set plausibly can be reduced to the mismatched data distributions, 
as Part-$A^2$ \cite{shi2020points2parts} notes. 

Overall, our BADet achieves decent performance on both \emph{val} set and \emph{test} set, which 
consistently demonstrates the effectiveness of our boundary-aware 3D detector.  

\subsection{3D Detection on the nuScenes Dataset}  \label{ResultsonthenuScenesDataset}
\highlight{
To further confirm the validity of the proposed method, we evaluate the performance of BADet 
on the highly challenging large-scale nuScenes Dataset.
\subsubsection{Evaluation Metric} We follow \emph{nuScenes Dataset} protocal by using 
mean Average Precision (mAP) and nuScenes detection score (NDS) as the main evaluation metrics for 
3D object detection.  The mAP assesses a bird-eye-view center offset within cut-off distances 
0.5m, 1m, 2m, 4m rather than a standard Intersection over Union (IoU). Let 
$E=\{ mATE,\ mASE,\ mAOE$ $,\ mAAE,\ mAVE \}$ be a set of 
the mean average errors of translation, size, orientation, attribute and velocity, 
the NDS denotes a weighted average metric, which is obtained by
\begin{equation}
	\begin{aligned}        
		NDS=\frac{1}{10}\left[ 5mAP+\sum_{err\in E}{\left( 1-\min \left( 1,\ err \right) \right)} \right].
		\label{eq:1}
	\end{aligned}      
\end{equation} 
% a typical long-tail distribution
\subsubsection{Comparison with State-of-the-Arts}
Table \ref{tab:APonnuScenesdataset} and Table \ref{tab:NDSonnuScenesdataset} reveal that our BADet outperforms 
3DSSD\cite{Yang2020ssd}\footnote{As of Nov.27,2021, 3DSSD is the state-of-the-art method among those manifested 
in Table \ref{tab:val11} } remarkably with 4.99\%, 2.44\% gains for mAP, NDS, respectively. Also, Table 
\ref{tab:APonnuScenesdataset} shows that BADet surpasses the existing state-of-the-art approaches remarkably
among most categories except for 3D $Car$, $Bus$, $Truck$, $Trailer$, $Construction\ vehicle$ detection, where the top three 
gains are 27.14\%, 18.47\%, 10.03\% for 3D $traffic\ cone$, $Bicycle$, $Pedestrian$ detection, 
which again validates the effectiveness of our region feature aggregation module in terms of 
aggregating informative contextual information, especially for small object detection. All results reported  
on the large-scale nuScenes Dataset further demonstrate the validity of the proposed method.
}

\vspace{1em}
\subsection{Ablation Study} \label{AblationStudies}
In this section, comprehensive ablation experiments are conducted to analyze the effectiveness 
of each component of our proposed BADet. Note that we report all results 
by \emph{Interpolated AP@$S_{11}$ Metric} with a IoU threshold 0.7 unless otherwise noted. 
We train all our models on the \emph{train} set and evaluate on the \emph{val} set for the most important 
\emph{Moderate} difficulty of \emph{Car} category  
of \emph{KITTI Dataset} \cite{geiger2012we,geiger2013vision}. 

\subsubsection{Effect of region feature aggregation} \label{Effectofregionfeatureaggregation}
\highlight{
To judiciously verify the effectiveness of our region feature aggregation module, 
we disable RFA module by directly adding a commonly used detect head on the top of the 2D backbone network in the SA-SSD \cite{he2020sassd}. 
As illustrated in the $2^{nd}$, $3^{rd}$ rows of Table \ref{tab:modules}, our RFA module 
typically integrates multi-grain semantic features in a coarse-to-fine manner and further improves the 
performance by (0.16\%, 0.13\%, 0.02\%) on all the three subsets.}

\subsubsection{Effect of different feature components for RFA module} \label{EffectofdifferentfeaturecomponentsforRFAmodule}
We closely verify each level feature component of RoI-wise representations in Eq. (\ref{eq:roiwisefeature}) by controlled experiments. As shown 
in the $1^{st}$, $2^{nd}$, and $3^{rd}$ rows of Table \ref{tab:fusion}, the performance of BADet deteriorates a lot 
when $\mathbf{f}^{\left( pixel \right)}$ from BEV representation is absent. Nevertheless, it can not achieve state-of-the-art 
performance by itself.   
\begin{table*}[htbp] %\tiny
	\centering
	\caption{
		\textbf{Performance of different models on nuScenes}, measured by the mean Average Precision (mAP). Numbers of SECOND, PointPillars and 3DSSD are cited from \cite{Yang2020ssd}.}
		% \vspace{-0.8em}
		\setlength{\tabcolsep}{1.5mm}{
	  \begin{tabular}{lccccccccccc}
	  \toprule
	  \textbf{Methods} & \textbf{Car} $\downarrow$   & \textbf{Ped} $\uparrow$  & \textbf{Bus} $\downarrow$   & \textbf{Barrier} $\uparrow$ & \textbf{TC} $\uparrow$   & \textbf{Truck} $\downarrow$ & \textbf{Trailer} $\downarrow$ & \textbf{Moto} $\uparrow$ & \textbf{Cons.Veh.} $\downarrow$ & \textbf{Bicycle} $\uparrow$ & \textbf{mAP} $\uparrow$ \\
	  \midrule
	  SECOND\cite{yan2018second} & 75.53 & 59.86 & 29.04 & 32.21 & 22.49 & 21.88 & 12.96 & 16.89 & 0.36  & 0     & 27.12 \\
	  PointPillar\cite{lang2019pointpillars} & 70.50 & 59.90 & 34.40 & 33.20 & 29.60 & 25.00 & 20.00 & 16.70 & 4.50  & 1.60  & 29.50 \\
	  3DSSD\cite{Yang2020ssd} & \textbf{81.20} & 70.17 & \textbf{61.41} & 47.94 & 31.06 & \textbf{47.15} & \textbf{30.45} & 35.96 & \textbf{12.64} & 8.63  & 42.66 \\ 
	  \emph{BADet} & 80.30 & \textbf{80.20} & 57.10 & \textbf{57.50} & \textbf{58.20} & 43.70 & 25.00 & \textbf{36.00} & 11.40  & \textbf{27.10} & \textbf{47.65} \\
	  \bottomrule
	  \end{tabular}%
	\label{tab:APonnuScenesdataset}}%
	\vspace{-1em}
\end{table*}%
\begin{table*}[htbp] %\scriptsize
	\centering
	\caption{\textbf{Performance of different models on nuScenes}, measured by the nuScenes Detection Score (NDS). Numbers of PointPillars and 3DSSD are cited from \cite{Yang2020ssd}.}
	% \vspace{-0.8em}  
	\begin{tabular}{@{}lccccccc@{}}
	  \toprule
			& \textbf{mAP} $\uparrow$   & \textbf{mATE} $\downarrow$  & \textbf{mASE} $\downarrow$ & \textbf{mAOE} $\downarrow$ & \textbf{mAVE} $\uparrow$ & \textbf{AAE} $\uparrow$  & \textbf{NDS} $\uparrow$  \\
	  \midrule
	  PointPillars \cite{lang2019pointpillars} & 29.50 & 0.54  & 0.29  & 0.45  & 0.29  & 0.41  & 44.90 \\
	  3DSSD \cite{Yang2020ssd} & 42.66 & 0.39  & 0.29  & 0.44  & \textbf{0.22} & \textbf{0.12} & 56.40 \\ [1pt]
	  %\midrule
	  %\midrule
	  \emph{BADet} & \textbf{47.65} & \textbf{0.30} & \textbf{0.27} & \textbf{0.34} & 0.41  & 0.18  & \textbf{58.84} \\
	  \bottomrule
	  \end{tabular}%
	\label{tab:NDSonnuScenesdataset}%
	\vspace{-1em}
  \end{table*}%
\begin{table*}[htbp] %\small
	\centering
	\caption{\textbf{Effect of region feature aggregation module and boundary-aware 3D detector.} Here, 
	``SA-SSD'', ``RFA'', ``BEV-Header'', and ``BA-Refiner'' denote the original architecture claimed in \cite{he2020sassd},
	region feature aggregation module, the commonly used detect head and our boundary-aware 3D detector with a detect head, respectively.}
	\label{tab:modules}%
	% \vspace{-0.6em}
	%\vspace{-1em}
	\setlength{\tabcolsep}{1.2mm}{
	  \begin{tabular}{cccccccccc}
	  \toprule
	  \multicolumn{4}{c}{\textbf{Components}}   & \multicolumn{3}{c}{$AP_{3D}\left( \% \right) $} & \multicolumn{3}{c}{$AP_{BEV}\left( \% \right) $} \\
	  SA-SSD & RFA & BEV-Header & BA-Refiner & \textbf{Easy}  & \textbf{Mod.} & \textbf{Hard}  & \textbf{Easy}  & \textbf{Mod.} & \textbf{Hard} \\
	  \midrule
	%   \midrule
	  \checkmark	&       &     &  & 90.04 & 79.78 & 78.91 & 90.64 & 89.12 & 80.49 \\
	  \checkmark	&       &  \checkmark  &   & 89.93 & 79.81 & 78.83 & 90.65 & 88.68 & 87.95 \\
	  \checkmark	&   \checkmark    &  \checkmark   &  & 90.09 & 79.94 & 78.85 & 90.66 & 88.76 & 88.05 \\
	  \checkmark	&   \checkmark    &    & \checkmark  & 90.06 & 85.77 & 79.00 & 90.63 & 88.86 & 88.10 \\
	  \bottomrule
	  \end{tabular}}%
	%   \vspace{-0.4em}
\end{table*}%
\begin{table*}[htbp] %\small
	\centering
	\caption{\textbf{Effect of each feature component of region feature aggregation module.}}
	% \vspace{-0.6em}
	\label{tab:fusion}%
	%\vspace{-1em}
	\setlength{\tabcolsep}{1.2mm}{
	  \begin{tabular}{ccccccccc}
	  \toprule
	  \multicolumn{3}{c}{\textbf{Components}}     & \multicolumn{3}{c}{$AP_{3D}\left( \% \right) $} & \multicolumn{3}{c}{$AP_{BEV}\left( \% \right) $} \\
	  $f^{\left( pixel \right)}$ & $f^{\left( voxel \right)}$ & $f^{\left( point \right)}$ & \textbf{Easy}  & \textbf{Mod.} & \textbf{Hard}  & \textbf{Easy}  & \textbf{Mod.} & \textbf{Hard} \\
	  \midrule
	%   \midrule
	  \checkmark     &       &       & 90.00 & 79.93 & 78.88 & 90.60 & 88.87 & 88.04 \\
	  & \checkmark   &       &  87.85	 & 76.24 & 72.96 & 89.60 & 82.83 & 78.73 \\	
	  &       &  \checkmark     &  88.36	 & 76.96 & 74.10 & 89.87 & 83.62 & 79.09 \\	
	       &  \checkmark     &  \checkmark     &  88.58	 & 77.88 & 75.37 & 89.93 & 84.65 & 79.48 \\	
	  \checkmark   &       &  \checkmark     &   90.07	 & 79.95 & 78.93 & 90.62 & 88.90 & 88.08 \\	
	  \checkmark    & \checkmark     &       & 89.99 & 80.00 & 78.97 & 90.57 & 88.96 & 88.07 \\
	  \checkmark    & \checkmark     & \checkmark    & 90.06 & 85.77 & 79.00 & 90.63 & 88.86 & 88.10 \\
	  \bottomrule
	  \end{tabular}}%
	%   \vspace{-0.4em}
\end{table*}%
This is because the progressive downsampled 3D feature volumes 
and the absence of 3D structure context when converting to BEV representation directly may degrade its localization 
accuracy. The aggregation of $\mathbf{f}^{\left( pixel \right)}$, $\mathbf{f}^{\left( voxel \right)}$, and $\mathbf{f}^{\left( point \right)}$ 
significantly benefits the performance as shown in the $5^{th}$ to $7^{th}$ rows. Note that each level feature component 
of $\mathbf{x}$ shows a supplementary effect more than a superposition. 
As shown in the last row, the aggregation of all feature components contributes to the top performance with 
85.77\% AP on the \emph{Moderate} difficulty of \emph{Car} category. 

\vspace{0.3em}
\subsubsection{Effect of boundary-aware 3D detector} \label{Effectofboundaryaware3Ddetector}
\vspace{0.3em}
\highlight{
As illustrated in the $3^{rd}$, $4^{th}$ rows of Table \ref{tab:modules}, our boundary-aware 3D detector 
raises the moderate AP (by 5.81\%), and hard AP (by 0.13\%) for KITTI car category 
of 3D detection and boosts the easy (by 0.11\%), moderate (by 0.12\%) and hard (0.08\%) categories 
of BEV detection without bells and whistles. The substantial improvements bring us a strong baseline, 
which validates that our proposed module could learn much richer boundary-aware contextual information 
in a coarse-to-fine manner by directly taking the semantic features from further reaches of its neighbors into account. 
}

\vspace{0.3em}
\subsubsection{Effect of boundary-aware 3D detector with T iterations} \label{Effectofboundaryaware3DdetectorwithTiterations} 
\vspace{0.3em}
We propose boundary-aware 3D detector 
in Sec. \ref{BoundaryAwareGraphNeuralNetwork} to model local boundary correlations of an object. Since graph neural network updates its nodes 
iteratively, the number of iterations is an important hyper-parameter. As shown in Table \ref{tab:iteration}, the 
$1^{st}$ row shows the nodes' state initialized by $\mathbf{x}$ can not achieve the top performance 
without our information compensation mechanism. Note that the performance is significantly improved as the receptive 
field is expanding via graph edges with $T$ iterations. We set $T$ to 3 for online test server submission. Whereas, 
for \emph{val} set, $T$=4 achieves the best performance. Particularly, when $T$ is set to 5, the performance drops 
dramatically, which is likely due to the problem of gradient vanishing. 
\begin{table}[htbp] %\small
	\centering
	\caption{\textbf{Ablation study on the KITTI val set with T iterations of BADet.}}
	\label{tab:iteration}%
	% \vspace{-0.6em}
	\setlength{\tabcolsep}{1.5mm}{
	  \begin{tabular}{ccccccc}
	  \toprule
	  \textbf{Number of}& \multicolumn{3}{c}{$AP_{3D}\left( \% \right) $} & \multicolumn{3}{c}{$AP_{BEV}\left( \% \right) $} \\
	  \textbf{iterations} & \textbf{Easy}  & \textbf{Mod.} & \textbf{Hard}  & \textbf{Easy}  & \textbf{Mod.} & \textbf{Hard} \\
	  \midrule
	%   \midrule
	  T=0   & 90.06 & 79.94 & 78.85 & 90.66 & 88.76 & 88.05 \\
	  T=1   & 85.68 & 80.16 & 77.84 & 88.72 & 88.13 & 87.76 \\
	  T=2   & 89.24 & 83.11 & 78.91 & 90.15 & 88.63 & 87.99 \\
	  T=3   & 90.06 & 85.77 & 79.00 & 90.63 & 88.86 & 88.10 \\
	  T=4   & 90.28 & 85.79 & 79.11 & 90.71 & 88.97 & 88.17 \\
	  \midrule
	  T=5   & 90.02 & 79.87 & 78.85 & 90.65 & 88.83 & 88.11 \\
	  \bottomrule
	  \end{tabular}}%
	%   \vspace{-0.4em}
\end{table}%
\begin{table}[htbp] %\footnotesize
	\centering
	\caption{\textbf{Runtime analysis for each component during inference stage. ms indicates millisecond.}}
	% \vspace{-0.6em}
	\setlength{\tabcolsep}{1.43mm}{
	  \begin{tabular}{ccccccc}
	  \toprule
	  \textbf{Components}  & \textbf{RPN}   & \textbf{RFA} & \textbf{BA-Refiner}  & \textbf{NMS}   & \textbf{Overall} \\
	  \midrule
	  Avg. time (ms) & 43.65 & 26.73  & 61.98 & 1.49  & 140.36 \\
	  \bottomrule
	  \end{tabular}}%
	\label{tab:runtime}%
	% \vspace{-1.9em}
\end{table}%
\subsection{Runtime Analysis} \label{RuntimeAnalysis}
The inference time is important for the deployment of downstream applications in the context of autonomous driving. 
Hence, we report a breakdown of runtime of our BADet for future optimization \ref{tab:runtime}. Our BADet is implemented by Pytorch 
with Python language. We set batch size to 1 and assess runtime on 
a single Intel(R) Core(TM) i7-6900K CPU and a single GTX 1080 Ti GPU. \highlight{
The average inference time (in millisecond) over \emph{val} set (3769 samples) of BADet is 140.36 ms, which is almost five 
times faster than Point-GNN \cite{Point-GNN} (7.1 FPS vs. 1.5 FPS): As show in Table \ref{tab:runtime}, 
(i) 3D proposal generation in the first stage takes 43.65ms (31.10\%); (ii)Region feature 
aggregation network takes 26.73ms (19.04\%); (iii) boundary-aware graph neural network takes 61.98ms (44.16\%); 
and non-maximum suppression for filtering redundancy takes 1.49ms (1.04\%). }In fact, as She et al. notes \cite{Point-GNN}, 
factors affecting the runtime varies from software level (\eg code optimization) to 
hardware level (\eg GPU resources). Optimizing runtime is out of the scope of this paper. However, an 
analysis of inference time facilitates more follow-up works in this field.

\vspace{-0.1em}
\section{Concluding Remarks}
\highlight{
We have presented BADet for 3D object detection from point clouds. Conceptually, the new model provides an information 
compensation mechanism to deal with the situation wherein RoI proposals lack enough information about the object 
boundary due to their deviations to the (unknown) ground truth. Extensive experiments on two public benchmarks, 
\ie the widely acknowledged KITTI and the more recent nuScenes, show that BADet compares favorably against the 
state-of-the-art. The encouraging results can set a new baseline for more follow-up literature and are poised to 
facilitate other downstream applications. }

% \highlight{
Much remains to be done. First, the inference time needs to be further optimized, say by down-scaling the 
\begin{table}[htbp] \footnotesize
	\centering
	\caption{\textbf{Comparing the runtime speed of our BADet with state-of-the-art detectors.}}
	% \vspace{-0.8em}
	%\vspace{-1em}
	\setlength{\tabcolsep}{0.1mm}{
	  \begin{tabular}{ccccccc}
	  \toprule
	  \textbf{Methods} & \textbf{PointRCNN} & \textbf{Part-}$A^2$ & \textbf{STD}  & \textbf{PV-RCNN} & \textbf{Point-GNN} & \textbf{BADet} \\
	  \midrule
	  Speed (FPS) & 10   & 12.5    & 12.5  & 12.5  & 1.5   & 7.1 \\
	  \bottomrule
	  \end{tabular}}%
	\label{tab:addlabel}%
	% \vspace{-0.8em}
  \end{table}%
local neighborhood graphs. Second, attention mechanisms 
can be integrated into the RFA module for adaptive multi-level 
feature aggregation. Finally, as we consider point clouds so far, how to extend BADet to handle multi-modality 
sensory data deserves further research. 
% }

\medskip  

\highlight{
\textbf{Acknowledgements}. This work was supported by the National Natural Science Foundation of China 
under Grant 62172420 and Beijing Natural Science Foundation under Grant 4202033.
}

\bibliographystyle{elsarticle-num}
% \bibliographystyle{cas-model2-names}
   
\vspace{1.1em}
\bibliography{references}

\end{document}